\documentclass{article}

\usepackage{microtype}
\usepackage{graphicx}
\usepackage{subcaption}
\usepackage{booktabs} 

\usepackage{hyperref}

\usepackage[preprint]{icml2026}

\usepackage{amsmath}
\usepackage{amssymb}
\usepackage{mathtools}
\usepackage{amsthm}
\usepackage{times}
\usepackage{soul}
\usepackage{url}
\usepackage{booktabs}
\usepackage{algorithm}
\usepackage{algorithmic}
\usepackage[switch]{lineno}
\usepackage{natbib}
\usepackage{amssymb}
\usepackage{xcolor}
\usepackage[frozencache=true,cachedir=minted-cache]{minted} 
\usepackage{multirow}
\usepackage{threeparttable}
\usepackage[capitalize,noabbrev]{cleveref}

\theoremstyle{plain}
\newtheorem{theorem}{Theorem}[section]
\newtheorem{proposition}[theorem]{Proposition}

\theoremstyle{definition}

\theoremstyle{remark}

\usepackage[textsize=tiny]{todonotes}

\icmltitlerunning{SWIFT}
\begin{document}

\twocolumn[
  \icmltitle{SWIFT: Mapping Sub-series with Wavelet Decomposition Improves \\ Time Series Forecasting}



  \icmlsetsymbol{equal}{*}

\begin{icmlauthorlist}
  \icmlauthor{Wenxuan Xie}{equal,scut}
  \icmlauthor{Fanpu Cao}{equal,scut}
\end{icmlauthorlist}
\icmlcorrespondingauthor{Wenxuan Xie}{lancelotxie601@gmail.com}
\icmlcorrespondingauthor{Fanpu Cao}{202164690237@mail.scut.edu.cn}
\icmlaffiliation{scut}{South China University of Technology, Guangzhou, China}

  \icmlkeywords{Machine Learning, ICML}

  \vskip 0.3in
]



\printAffiliationsAndNotice{\textsuperscript{*}Equal contribution. }


\begin{abstract}
In recent work on time-series prediction, Transformers and even large language models have garnered significant attention due to their strong capabilities in sequence modeling. However, in practical deployments, time-series prediction often requires operation in resource-constrained environments, such as edge devices, which are unable to handle the computational overhead of large models. To address such scenarios, some lightweight models have been proposed, but they exhibit poor performance on non-stationary sequences. In this paper, we propose \textit{SWIFT}, a lightweight model that is not only powerful, but also efficient in deployment and inference for Long-term Time Series Forecasting (LTSF). Our model is based on three key points: (\romannumeral 1) Utilizing wavelet transform to perform lossless downsampling of time series. (\romannumeral 2) Achieving cross-band information fusion with a learnable filter. (\romannumeral 3) Using only one shared linear layer or one shallow MLP for sub-series' mapping. We conduct comprehensive experiments, and the results show that \textit{SWIFT} achieves state-of-the-art (SOTA) performance on multiple datasets, offering a promising method for edge computing and deployment in this task. Moreover, it is noteworthy that the number of parameters in \textit{SWIFT-Linear} is only 25\% of what it would be with a single-layer linear model for prediction.  

\end{abstract}

\section{Introduction}

\begin{figure*}[ht]
  \centering
  \includegraphics[width=\textwidth]{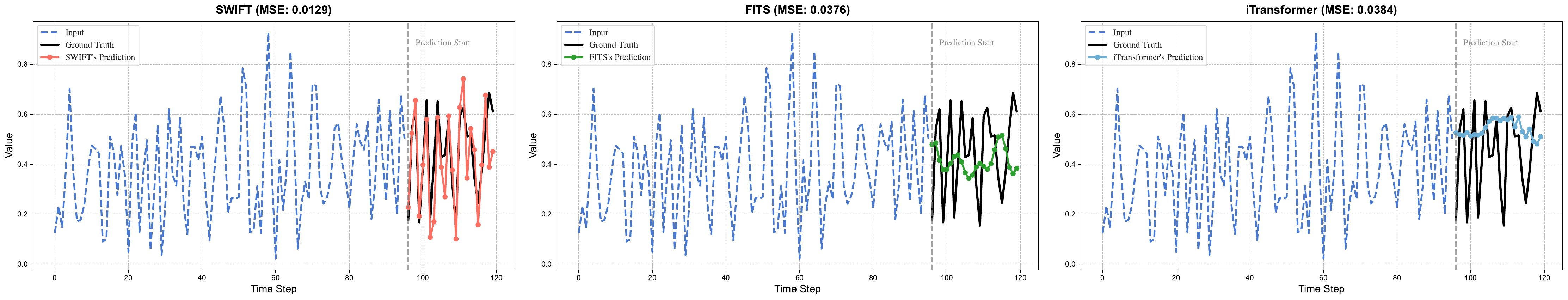}
  \caption{MSE performance on a simple synthetic non-stationary signal, with forecasting starting at the 96th time step.}
  \label{fig:sample}
\end{figure*}

Long-term time series forecasting (LTSF) finds broad applications in various domains, including energy management, financial market analysis, weather prediction, traffic flow monitoring, and healthcare monitoring. Accurate prediction is crucial for them. At the same time, many applications require real-time prediction on edge devices, e.g., in latency-sensitive tasks such as energy scheduling or intelligent transportation systems, where models need to be responsive to real-world demands and where edge computing and fast inference are critical. Additional challenges are posed under the conditions of limited computational resources \citep{bigger}.

Achieving precise forecasts typically relies on powerful yet complex deep learning models, such as RNNs \citep{rnn}, LSTMs \citep{lstm}, TCNs \citep{tcn, timesnet}, and Transformers \citep{informer, autoformer, fedformer}. Thanks to the self-attention mechanism, Transformers can capture long-range dependencies in sequences, which improves prediction accuracy and makes them the most powerful of the existing LTSF forecasters. Additionally, LLMs’ impressive capabilities have inspired their application to time series forecasting. Many LLM-based LSTF forecasters are proposed \citep{time-llm, gpt4ts}. However, these models also face several challenges stemming from their \textit{computational inefffciency} and \textit{the large scale of their model weights}, which restrict their practical applicability, particularly in environments with limited computational resources. Since recently a solid paper \citep{DLinear} has shown that even a simple one-layer linear model can outperform transformer-based models in almost all cases, more and more efficient linear forecasters are proposed \citep{tide, koopa, FITS}. While improving prediction accuracy, these linear forecasters are constantly becoming more efficient, with faster inference speed and less deployment costs, pushing the boundary of this field forward. Recently, FITS \citep{FITS} modeling time-series with a complex-valued neural network, surpassed several existing Transformer models in both inference speed and forecasting performance with $10k$ parameters, establishing itself as a benchmark in the field. 

However, existing linear-based models often suffer from suboptimal performance when dealing with non-stationary time series, and in some cases, they fail entirely to fit the ground truth. Through an analysis of state-of-the-art Transformer-based and linear-based models, we found that they struggle to achieve accurate predictions on a simple synthetic non-stationary dataset (Figure \ref{fig:sample}). This indicates that current linear models are insufficient for handling non-stationary sequences, which are prevalent in real-world scenarios and industrial applications, posing significant challenges for the deployment of such models. There is an urgent need to develop an efficient time series model capable of effectively handling non-stationary data. Motivated by the above observations, we present \textit{SWIFT}, a lightweight model based on first order wavelet transform \citep{multiwaveleet} and only one linear layer. \textit{SWIFT} achieves strong performance, while being at least $100\times$ lighter than several mainstream LTSF models (e.g., PatchTST\citep{patch}).
 Even when compared to a lightweight model like FITS, our model still has only 15\% of its number of parameters. 

In summary, our contributions can be delineated as follows: 
\begin{itemize}  
  \item In studying state-of-the-art LTSF forecasters, we found that existing models are either difficult to deploy under limited resource constraints or incapable of accurately handling non-stationary sequence forecasting tasks. This motivated us to develop an efficient time series model capable of effectively addressing non-stationary data.
  \item We propose \textit{SWIFT}, a powerful lightweight model for time-series forecasting tasks, which is four times smaller than the single-layer linear model for time-domain prediction. We employ wavelet transform as a nearly lossless downsampling method, which is the key to SWIFT's ability to maintain good performance while significantly reducing the number of parameters.
  \item We conduct extensive experiments on predicting long multivariate sequences on several real-world benchmarks showing the superiority of our method in terms of effectiveness and efffciency.
\end{itemize}

\section{Related Work}

\subsection{Efficient linear forecasters} Since \citep{DLinear} shows that a simple one-layer linear model can outperforms Transformer forecasters \citep{informer, autoformer, fedformer} in almost all cases, there has been a rapid emergence of linear forecasters \citep{N-Beats, tide, koopa} in LTSF. The impressive performance and efficiency continuously challenge this direction. \cite{cyclenet} employs a Linear backbone combined with learnable recurrent cycles to explicitly model periodic structures in time series. Recently, FITS \citep{FITS} introduced a frequency-domain interpolation strategy that utilizes low-pass filters and FFT \citep{fft} for time series modeling. With $10k$ parameter, FITS surpassed several existing Transformer models in both inference speed and forecasting performance, establishing itself as a benchmark in the field.

However, FFT assumes that signals are stationary, limiting its ability to capture the temporal localization of transient or non-stationary signals \citep{Non-station}. Meanwhile, the limited representational capacity of a single-layer linear model typically necessitates a longer look-back window to prevent underfitting and distribution shifts. The parameter count of FITS is primarily determined by the length of the look-back window due to its interpolation-based prediction approach. Consequently, The efficiency of FITS decreases significantly as the lookback window increases. Our proposed model, SWIFT, aims to enhance the field of efficient time series forecasting through the introduction of DWT. This approach not only improves SWIFT's capacity to handle non-stationary signals but also significantly reduces model's parameter count, thereby enhancing efficiency while preserving predictive performance.

\begin{figure*}[t]
  \centering
  \includegraphics[width=1.0\textwidth]{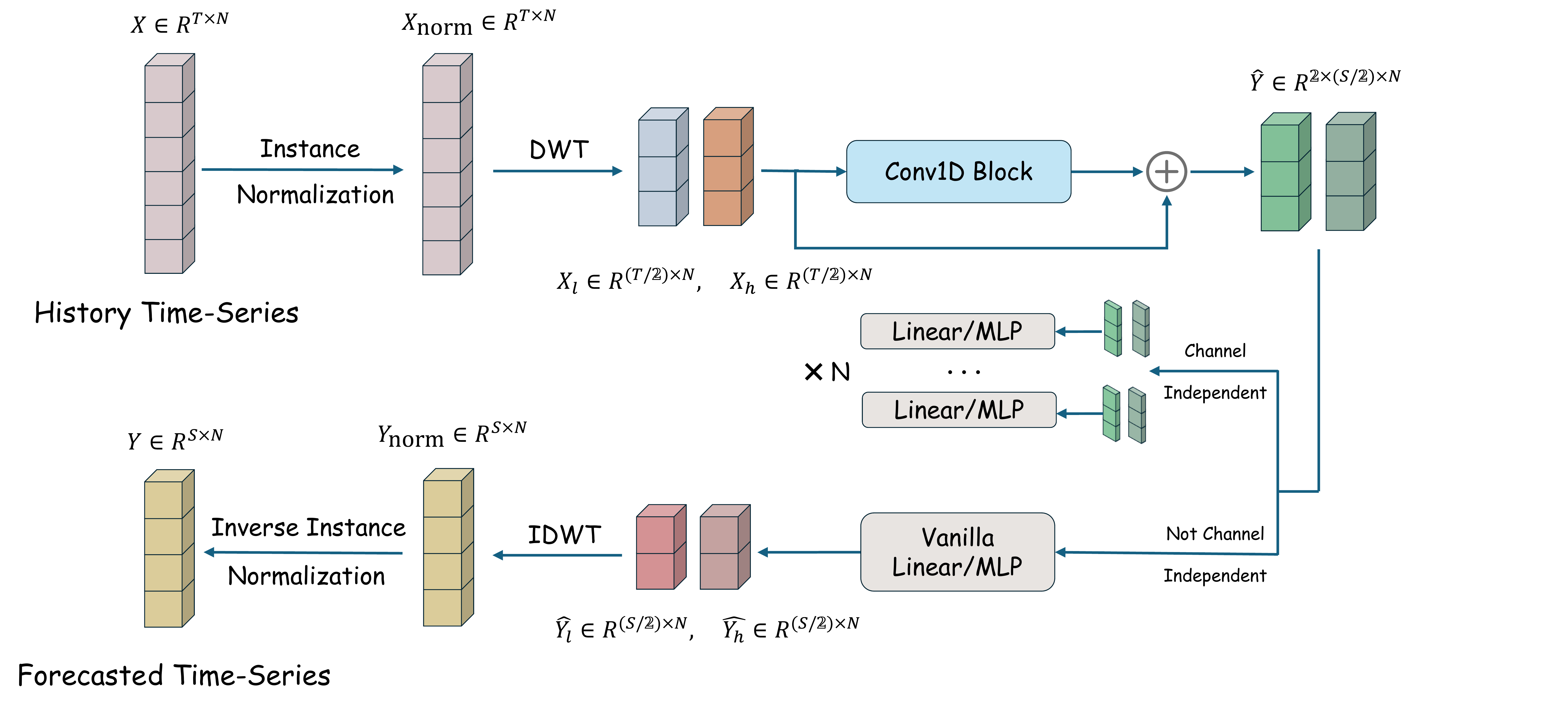}
  \caption{Overall structure of \textit{SWIFT}. (\romannumeral 1)The DWT module decomposes the time series into two sub-series: the approximation coefficient and the detail coefficient, based on the Haar wavelet; (\romannumeral 2) The convolutional layer is applied for filtering and feature aggregation. (\romannumeral 3)Linear or MLP is used for the mapping of sub-series to make prediction. \textit{T} denote the length of the look-back window, \textit{N} is the number of variables (i.e., channels), and \textit{S} refers to the length of the prediction horizon.}
  \label{fig:model}
\end{figure*}

\subsection{DWT method}
As a powerful method for time-frequency analysis, DWT is widely used in tasks dealing with time series. \citep{Wavelet-CNN-LSTM} decomposed the time-series using wavelet decomposition and then utilized CNN and LSTM for prediction. \citep{fedformer} combines Wavelet Transform with frequency enhanced strategy and attention mechanism to capture long range dependencies. \citep{W-Transformers} utilize a maximal overlap discrete wavelet transformation and build a local transformer model for time-series forecasting. \citep{adawavenet} utilizes learnable lifting-based wavelet transforms to adaptively model non-stationary time series. \citep{wpmixer} leverages level-specific wavelet coefficient decomposition combined with patch-based mixing mechanisms to preserve multi-resolution information. DWT is also often used for anomaly detection. For instance, \citep{Multi-damage-detection} used wavelet transform for signal denoising and damage localization. Recently, \citep{wavemask} proposed a method for data augmentation in time-series prediction tasks, which is used to obtain more diverse sequences by eliminating or swapping wavelet coefficients. However, none of these approaches provided significant insight into the wavelet transform for LTSF. In our work, we explored the wavelet coefficients in depth and found that the high-frequency coefficients and low-frequency coefficients of the historical wavelet can be mapped to the coefficients of the future wavelet in the same representation space.

\section{Preliminary}

\subsection{LTSF problem deﬁnition}
In multivariate LTSF, time series data contain multiple variables at each time step. Given historical values $\mathcal{X}=\{\mathbf{x}_1,\ldots,\mathbf{x}_{L_x} ~|~ \mathbf{x}_{i} \in \mathbb{R}^{d}\}$ where $d$ represents the number of variables and $L_x$ represents the length of the look-back window, the objective of LTSF is to predict future values $\mathcal{Y}=\{\mathbf{y}_1,\ldots,\mathbf{y}_{L_y} ~|~ \mathbf{y}_{i} \in \mathbb{R}^{d} \}$. Notably, the feature dimension is not limited to the univariate case ($d \geq 1$).

\subsection{DWT and time-frequency domain}
With the discrete wavelet transform, the signal is decomposed into a series of linear combinations of wavelet functions and scale functions, the coefficients of each combination being the wavelet coefficients. In DWT, the scale function $\phi(t)$ and wavelet function $\psi(t)$ are related as follows:
\begin{equation}  
\begin{aligned}
\phi(t) & =\sum_{n} h_{\phi}[n] \sqrt{2} \phi(2 t-n) \\
\psi(t) & =\sum_{n} h_{\psi}[n] \sqrt{2} \phi(2 t-n)
\end{aligned}
\end{equation}
At the same time, we are able to obtain recursive formulas for approximation coefficients $W_{\phi}[j, k]$ and detail coefficients $W_{\psi}[j, k]$, where j denotes the order of the wavelet decomposition and k denotes the shift in the time domain. 
\begin{equation} 
\begin{aligned}
&W_{\phi}[j, k]=h_{\phi}[-n] * W_{\phi}[j+1, n] \\
&W_{\psi}[j, k]=h_{\psi}[-n] * W_{\phi}[j+1, n]
\end{aligned}
\end{equation}

In SWIFT, We choose the Haar wavelet and perform only the first order decomposition (j = 1), and its filters corresponding to the scale and wavelet functions are:
\begin{equation}  
\begin{aligned}
&h_{\phi}[n] = \{1 / \sqrt{2}, 1 / \sqrt{2}\} \\
&h_{\psi}[n] = \{1 / \sqrt{2},-1 / \sqrt{2}\}
\end{aligned}
\end{equation}
We select Haar because it can make the transform fast and stable, which increases the speed of reasoning across our framework. Although higher-order wavelets such as Daubechies and Symlet generally exhibit better smoothness and frequency localization properties, the Haar wavelet excels at capturing sharp transitions and local discontinuities, which are often critical in time series forecasting tasks involving abrupt changes or short-term pattern shifts. Furthermore, in our implementation of SWIFT, the use of a $1_{st}$-order Haar decomposition (length=2) offers the added advantage of mitigating edge effects in discrete wavelet transforms (DWT) by requiring minimal boundary data. These considerations collectively make the Haar basis particularly well-suited for our application.

\section{Proposed Method}
\subsection{Structure Overview}
We propose the \textit{SWIFT} which is shown in Figure \ref{fig:model}. Firstly, time-series is decomposed by $1_{st}$ order DWT. Then the high-frequency component and the low-frequency component are concatenated and mapped using the same backbone layer after passing through a learnable filter. Finally, the prediction is obtained by performing IDWT on the new components obtained from the mapping.

\subsection{SWIFT Components}

\paragraph{Instance Normalization} Distribution shifts between training and testing datasets are common in time series data. Recent studies \citep{Non-station} have shown that applying simple instance normalization strategies between the model input and output can effectively mitigate this issue. We also adopt a straightforward normalization approach in SWIFT. Specifically, before feeding the sequence into the model, we subtract its mean value. After the model produces its output, we add the mean value back to reconstruct the original scale. This process can be formally expressed as follows:  
\[ \tilde{\mathcal{X}} = \mathcal{X} - \bar{\mathcal{X}}, \]
\[ \hat{\mathcal{Y}} = f(\tilde{\mathcal{X}}) + \bar{\mathcal{X}}, \]

where \( \mathcal{X} \) represents the input sequence, \( \tilde{\mathcal{X}} \) is the normalized sequence, \( \bar{\mathcal{X}} \) is the mean value of the sequence, \( f(\cdot) \) is the model, and \( \hat{\mathcal{Y}} \) is the reconstructed output.

Meanwhile, due to the different distributional bias phenomena existing in different datasets, we used two instance norm approaches. We take the norm approach in ReVIN \citep{Revin} and use a dynamically learnable regularized representation to better combat distributional bias. 

\[
\mathcal{X}=\gamma\left( \frac{\mathcal{X}-\mathbb{E}_{t}\left[\mathcal{X}\right]}{\sqrt{\operatorname{Var}\left[\mathcal{X}\right]+\epsilon}}\right)+\beta
\]

\paragraph{Channel-Independence} Channel-Independence (CI) is a strategy that simplifies multivariate time series forecasting by focusing on individual univariate sequences within the dataset. Many advanced forecasters employ it, such as DLinear \citep{DLinear}, PatchTST \citep{patch} and TiDE \citep{tide}. Instead of modeling interdependencies between channels, CI treats each channel independently, reducing the overall complexity of the prediction task. Specifically, the CI approach learns a shared function for mapping historical univariate data to future predictions, effectively decoupling the relationships between channels. This independence enables the model to focus on intra-channel patterns like trends and periodicity without being affected by inter-channel noise or variability. Furthermore, CI can significantly enhance the scalability and generalization of forecasting models, especially when handling datasets with numerous channels. 

In designing SWIFT, we adopt the CI strategy to harness its benefits for capturing long-term dependencies while maintaining model simplicity and efficiency. By leveraging CI, SWIFT reduces computational overhead and achieves robust performance across diverse time series forecasting tasks.

\paragraph{DWT decomposition} Given an input sequence $\mathbf{X} \in \mathbb{R}^{N \times T}$, where $T$ is the length of lookback window, and $N$ is the number of variables, we apply a single-level DWT based on Haar wavelet:

\begin{equation}
\mathcal{Y_L}, \mathcal{Y_H} = \text{DWT}(\mathbf{X})
\end{equation}

where $\mathcal{Y_L} \in \mathbb{R}^{N \times T/2}$ represents the approximation coefficients (low-frequency components), and $\mathcal{Y_H} \in \mathbb{R}^{N \times T/2}$ represents the detail coefficients (high-frequency components). The low-frequency component, obtained by convolving the input signal with a low-pass filter, captures the overall trend and smooth variations in the original signal. The high-frequency component, obtained by convolving the input signal with a high-pass filter, represents the rapid variations, discontinuities, and fine-scale structures in the signal. The length of each segment of coefficients is only half of the original time series, achieving a nearly lossless downsampling process.

We employ a novel sub-series mapping strategy that leverages the multi-resolution analysis capabilities of DWT. This approach allows us to capture and project both low-frequency trends and high-frequency details of the input time series efficiently. Our key innovation lies in the unified mapping of both low and high-frequency components. We concatenate these two components along a new dimension to gain the time-frequency representation of whole series:

\begin{equation}
\mathbf{Y} = \mathcal{[Y_L;Y_H]} \in \mathbb{R}^{N \times 2 \times T/2}
\end{equation}

After obtaining the representation \(\mathbf{Y}\), SWIFT extracts information from this representation by means of convolution and Mapping.

\paragraph{Learnable filter for aggregating and filtering}
In our experiments, it has been found that there is some commonality in the different band coefficients, with the potential to map through the same representation space. In addition, the timing characteristics of the coefficient vector need to further aggregated. Light weight convolutional layers has the ability to efficiently extract temporal features. They are the ideal solution to this problem. Therefore, we add a 1D convolutional layer with input channel 2 and output channel 2. By presetting the kernel size and stride length, the sequence length before and after convolution remains constant.

After the aggregation of information, we are able to get new components:
\begin{equation}
\mathbf{Y_C} = Conv(\mathbf{Y}) + \mathbf{Y}
\end{equation}

In SWIFT, the convolution layer serves three primary functions: (i) denoising the signal, enhancing features, and acting as an effective filtering mechanism; (ii) aggregating local information to capture long- and short-term dependencies in time series; and (iii) enabling cross-band information fusion, allowing coefficients from different bands to share a linear layer.

In time series forecasting, the receptive field is crucial. A larger receptive field allows the model to capture extensive temporal dependencies, improving prediction accuracy. Our proposed SWIFT model, which uses wavelet-domain convolution, offers significant advantages in receptive field expansion.

Traditional convolutional methods struggle to expand the receptive field. Increasing kernel size quadratically increases parameters, causing over-parameterization and computational inefficiency. Additionally, performance gains saturate before achieving a global receptive field. In contrast, SWIFT's wavelet convolution provides an elegant alternative.

This approach achieves a larger effective receptive field with minimal growth in trainable parameters. Cascading wavelet decomposition increases frequency resolution at each level while expanding the receptive field. For example, with an $\ell$-level cascading frequency decomposition and a fixed kernel size $k$, the receptive field grows exponentially as $2^{\ell} \cdot k$, while parameters scale linearly as $\ell \cdot 2 \cdot k^{2}$ (where $c$ is the number of channels). In contrast, traditional methods exhibit quadratic parameter growth relative to receptive field size.

\begin{figure}[ht]
  \centering
  \includegraphics[width=0.48\textwidth]{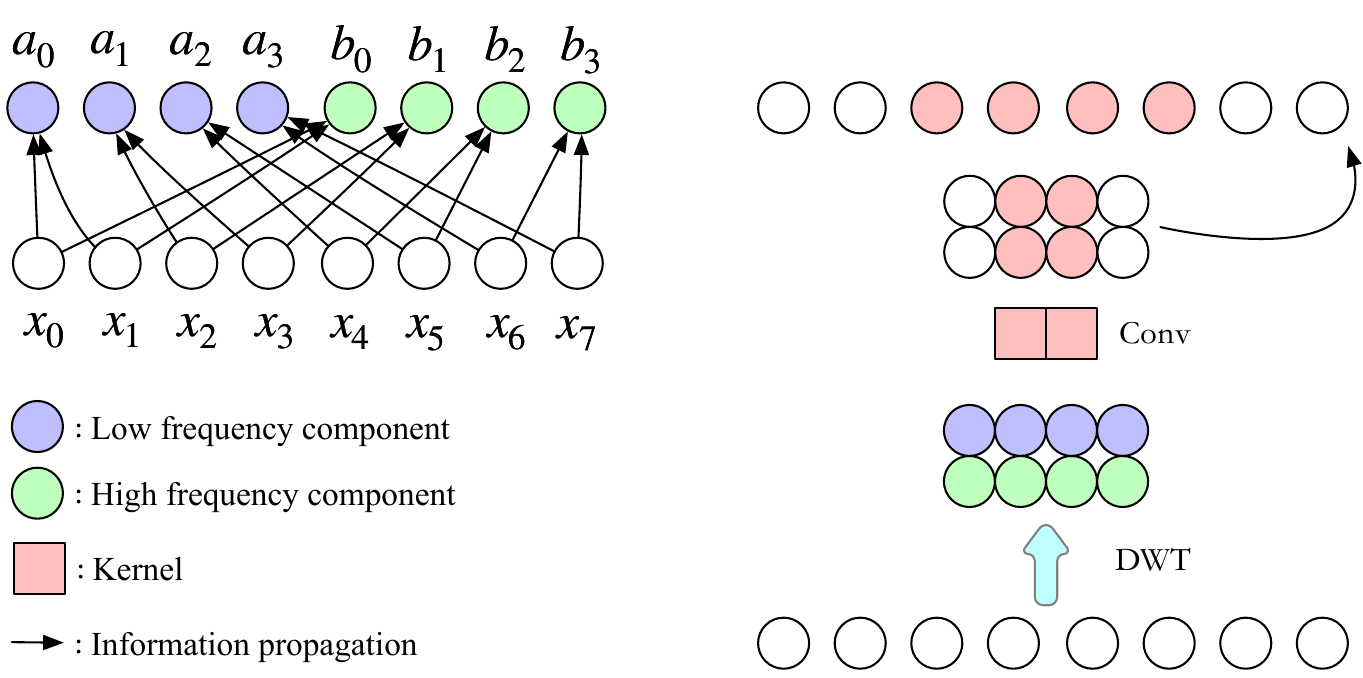}
  \caption{Performing convolution in the wavelet domain ($\ell=1$) results in a larger receptive field. In this example, a convolution is able to have a receptive field of 4 with a kernel size of 2.}
  \label{fig:conv}
\end{figure}

\paragraph{Sub-series mapping strategy} 
As mentioned in the previous sections, we use DWT to handle the time series and divide it into two sub-series ($\mathcal{Y_L}$ and $\mathcal{Y_H}$) by $1st$ order decomposition. The obtained components are concatenated as the time-frequency representation of whole sequence $\mathbf{Y}$. In SWIFT, we use Linear or MLP to map the sub-series. 

Single-layer linear or MLP models, despite their widespread use, are constrained by inherent limitations in their representational capacity. These limitations often manifest as underfitting or overfitting phenomena, particularly when applied to complex, non-stationary time series data. Such models are susceptible to being disproportionately influenced by specific patterns within the data, potentially leading to degraded predictive performance.

To address these challenges and enhance the robustness of the mapping module, while simultaneously reducing the model's parameter count and improving inference speed, we propose a novel mapping strategy. Here we take a single Linear layer as an example. Our approach employs a shared weight matrix for mapping both low-frequency and high-frequency components of the input series:

\begin{equation}  
\mathbf{Y'} = \mathbf{Y_CW + b}
\end{equation}

where $\mathbf{W} \in \mathbb{R}^{T/2 \times T'/2}$ is the weight matrix, and $\mathbf{b} \in \mathbb{R}^{T'/2}$ is the bias vector. The resulting $\mathbf{Y'} \in \mathbb{R}^{N \times 2 \times T'/2}$, where $T'$ is the prediction length. After mapping, we reshape $\mathbf{Y'}$ back into approximation and detail coefficients, and apply the Inverse Discrete Wavelet Transform (IDWT) to obtain the final prediction:

\begin{equation}  
\begin{aligned}
\mathbf{Y'_L = Y'_{:,0,:}}, \mathbf{Y'_H = Y'_{:,1,:}} \\
\mathbf{\hat{Y}} = \text{IDWT}(\mathbf{Y'_L, Y'_H}), \mathbf{\hat{Y}} \in \mathbb{R}^{N \times T'}
\end{aligned}
\end{equation}

The shared mapping strategy has several advantages. (1) It enhances the robustness of the model by jointly handling the low and high frequency components, reduces the sensitivity to the presence of a single specific pattern in the time series, and mitigates the occurrence of overfitting. The shared weights encourage the model to learn generalized features applicable to both frequency ranges, thus enhancing the robustness of the model. (2) It improves parameter efficiency by using a single Linear backbone for both components, which significantly reduces the total number of parameters, improves the computational efficiency of the model and speeds up inference time compared to mapping the low-frequency and high-frequency components separately. Together, these advantages enhance the prediction performance in time series forecasting involving complex, non-stationary data.

It is worth noting that in our experiments, we found that using an MLP to perform the mapping achieved excellent results on multivariate datasets (traffic, electricity). This is because the influence of multiple variables can be well-fitted by the MLP when the lookback window length is sufficiently large. Therefore, even when predicting based on the channel independence strategy, the MLP can still learn the influence of other variables. The proof of this part is provided in the appendix.

\section{Experiment}

\begin{table*}[ht]
\centering
\resizebox{\textwidth}{!}{
\begin{tabular}{@{}c|ccccc|ccccc|ccccc|ccccc@{}}
\toprule
Dataset & \multicolumn{5}{c|}{ETTh1} & \multicolumn{5}{c|}{ETTh2} & \multicolumn{5}{c|}{ETTm1} & \multicolumn{5}{c}{ETTm2} \\ \midrule
Horizon & 96 & 192 & 336 & 720 & Avg & 96 & 192 & 336 & 720 & Avg & 96 & 192 & 336 & 720 & Avg & 96 & 192 & 336 & 720 & Avg \\ 
\midrule
FEDFormer & 0.375 & 0.427 & 0.459 & 0.484 & 0.436 & 0.340 & 0.433 & 0.508 & 0.480 & 0.440 & 0.362 & 0.393 & 0.442 & 0.483 &0.420 & 0.189 & 0.256 & 0.326 & 0.437 & 0.302\\
TimesNet & 0.384 & 0.436 & 0.491 & 0.521 & 0.458 & 0.340 & 0.402 & 0.452 & 0.462 & 0.414 & 0.338 & 0.374 & 0.410 & 0.478 & 0.400 & 0.187 & 0.249 & 0.321 & 0.408 & 0.291\\ 
Dlinear & 0.384 & 0.443 & 0.446 & 0.504 & 0.444 & 0.282 & 0.350 & 0.414 & 0.588 & 0.409 & \underline{0.301} & \underline{0.335} & 0.371 & 0.426 & 0.358 & 0.171 & 0.237 & 0.294 & 0.426 & 0.282\\
PatchTST & 0.385 & 0.413 & 0.440 & 0.456 & 0.424 & 0.274 & 0.338 & 0.367 & 0.391 & 0.343 & \textbf{0.292} & \textbf{0.330} & 0.365 & 0.419 & \textbf{0.352} & 0.163 & 0.219 & 0.276 & 0.368 & 0.257 \\
iTransformer & 0.386 & 0.441 & 0.487 & 0.503 & 0.454 & 0.297 & 0.380 & 0.428 & 0.427 & 0.383 & 0.334 & 0.377 & 0.426 & 0.491 & 0.407 & 0.180 & 0.250 & 0.311 & 0.412 & 0.288 \\
FITS & \underline{0.372} & \underline{0.404} & \underline{0.427} & \textbf{0.424} & \underline{0.407} & \underline{0.271} & \underline{0.331} & \underline{0.354} & \textbf{0.377} & \underline{0.333} & 0.303 & 0.337 & \underline{0.366} & \underline{0.415} & \underline{0.355} & 0.162 & \underline{0.216} & \underline{0.268} & \textbf{0.348} & \underline{0.249} \\ 
CycleNet & 0.379 & 0.416 & 0.447 & 0.477 & 0.430 & \underline{0.271} & 0.332 & 0.362 & 0.415 & 0.345 & 0.307 & 0.337 & 0.364 & 0.410 & \underline{0.355} & \textbf{0.159} & \textbf{0.214} & \underline{0.268} & \underline{0.353} & \underline{0.249}\\
\midrule
SWIFT / MLP & 0.383 & 0.439 & 0.469 & 0.476 & 0.442 & 0.305 & 0.349 & 0.372 & 0.416 & 0.361 & 0.305 & 0.330 & 0.368 & 0.444 & 0.362 & 0.170 & 0.233 & 0.278 & 0.355 & 0.259 \\ 
SWIFT / Linear & \textbf{0.367} & \textbf{0.395} & \textbf{0.420} & \underline{0.430} & \textbf{0.403} & \textbf{0.268} & \textbf{0.329} & \textbf{0.351} & \underline{0.383} & \textbf{0.333} & 0.307 & 0.336 & \textbf{0.364} & \textbf{0.413} & \underline{0.355}& \underline{0.161} & \textbf{0.214} & \textbf{0.267} & \textbf{0.348} &\textbf{0.248} \\ 
\midrule
STD & 0.000 & 0.000 & 0.000 & 0.000 & 0.000 & 0.000 & 0.000 & 0.000 & 0.000 & 0.000 & 0.000 & 0.000 & 0.000 & 0.000 & 0.000 & 0.000 & 0.000 & 0.000 & 0.000 & 0.000\\
IMP & 0.005 & 0.009 & 0.007 & -0.006 & 0.004 & 0.003 & 0.002 & 0.003 & -0.006 & 0.000 & -0.015 & -0.006 & 0.002 & 0.002 & -0.003 & -0.002 & 0.000 & 0.001 & 0.000 & 0.001 \\
\bottomrule
\end{tabular}
}
\caption{Long-term forecasting results on ETT dataset in MSE. The best result is highlighted in \textbf{bold}, and the second best is highlighted with \underline{underline}. IMP is the improvement between SWIFT and the best baseline models, where a larger value indicates a better improvement. Most of the STD are under 5e-4 and shown as 0.000 in this table. }
\label{tab:SWIFTett}
\end{table*}

\begin{table*}[ht]
\centering
\resizebox{0.85\textwidth}{!}{
\begin{tabular}{@{}c|ccccc|ccccc|ccccc@{}}
\toprule
Dataset & \multicolumn{5}{c|}{Weather} & \multicolumn{5}{c|}{Electricity} & \multicolumn{5}{c}{Traffic} \\ 
\midrule
Horizon & 96 & 192 & 336 & 720 & Avg & 96 & 192 & 336 & 720 & Avg & 96 & 192 & 336 & 720 & Avg\\ 
\midrule
FEDformer & 0.246 & 0.292 & 0.378 & 0.447 & 0.341 & 0.188 & 0.197 & 0.212 & 0.244 & 0.210 & 0.573 & 0.611 & 0.621 & 0.630 & 0.609\\
TimesNet & 0.172 & 0.219 & 0.280 & 0.365 & 0.259 & 0.168 & 0.184 & 0.198 & 0.220 & 0.193 & 0.593 & 0.617 & 0.629 & 0.640 & 0.620 \\
Dlinear & 0.174 & 0.217 & 0.262 & 0.332 & 0.246 & 0.140 & 0.153 & 0.169 & 0.204 & 0.167 & 0.413 & 0.423 & 0.437 & 0.466 & 0.435\\
PatchTST & {0.151} & {0.195} & {0.249} & {0.321} & 0.229 & \underline{0.129} & 0.149 & 0.166 & 0.210 & 0.164 & {\textbf{0.366}} & \underline{0.388} & \underline{0.398} & 0.457 & 0.402 \\
iTransformer & 0.174 & 0.221 & 0.278 & 0.358 & 0.258 & 0.148 & 0.162 & 0.178 & 0.225 & 0.178 & 0.395 & 0.417 & 0.433 & 0.467 & 0.428 \\
FITS & \underline{0.143} & \underline{0.186} & \underline{0.236} & \textbf{0.307} & \underline{0.218} & {0.134} & 0.149 & 0.165 & 0.203 & 0.163 & 0.385 & 0.397 & 0.410 & \underline{0.448} & 0.410 \\
CycleNet & 0.149 & 0.192 & 0.242 & \underline{0.312} & 0.224 & \textbf{0.127} & \textbf{0.144} & \textbf{0.159} & \textbf{0.196} & \textbf{0.157} & 0.374 & 0.390 & 0.405 & 0.441 & \underline{0.403} \\
\midrule
 SWIFT / Linear & 0.159 & 0.201 & 0.243 & 0.325 & 0.232 & 0.133 & \underline{0.148} & 0.164 & 0.203 & \underline{0.162} & 0.385 & 0.396 & 0.410 & 0.448 & 0.410\\
SWIFT / MLP & \textbf{0.140} & \textbf{0.183} & \textbf{0.235} & \textbf{0.307} & \textbf{0.216}  & \textbf{0.127} & \textbf{0.144} & \underline{0.160} & \underline{0.197} & \textbf{0.157} & \underline{0.368} & \textbf{0.382} & \textbf{0.396} & \textbf{0.430}  & \textbf{0.394}\\
\midrule
STD & 0.000 & 0.000 & 0.000 & 0.000 & 0.000 & 0.000 & 0.000 & 0.000 & 0.000 & 0.000 & 0.000 & 0.000 & 0.000 & 0.000 & 0.000\\
IMP & 0.003 & 0.003 & 0.001 & 0.000 & 0.002 & 0.000 & 0.000 & -0.001 & -0.001 & 0.000 & -0.002 & 0006 & 0.002 & 0.018 & 0.009 \\ 
\bottomrule
\end{tabular}
}
\caption{Long-term forecasting results on three popular datasets in MSE. The best result is highlighted in \textbf{bold} and the second best is highlighted with \underline{underline}. IMP is the improvement between SWIFT and the best baseline models, where a larger value indicates a better improvement. Most of the STD are under 5e-4 and shown as 0.000 in this table. }
\label{tab:SWIFTother}
\end{table*}

\subsection{Forecasting results}

Our proposed model framework aims to improve performance and efficiency in LTSF, and we thoroughly evaluate SWIFT on various time series forecasting applications.

\paragraph{Datasets} We extensively include 7 real-world datasets in our experiments, including, Traffic, Electricity, Weather, ETT (4 subsets) used by Autoformer \citep{autoformer}. We summarize the characteristics of these datasets in appendix. 

\paragraph{Baselines} We carefully choose well-acknowledged forecasting models as our benchmark, including (1) Transformer-based methods: FEDformer \citep{fedformer}, PatchTST \citep{patch} and iTransformer \citep{itransformer}. (2) Efficient Linear-based methods: DLinear \citep{DLinear}, FITS \citep{FITS}, CycleNet \citep{cyclenet}. (3) TCN-based methods: TimesNet \citep{timesnet}. 

\paragraph{Implementation details} Our method is trained with the ADAM optimizer \citep{adam}, using OneCycleLR strategy \citep{pytorch} to adjust the learning rate. For the kernel size of the filter, we choose suitable size in \{3, 9, 13, 17\}. We evaluate all models across prediction horizons of \{96, 192, 336, 720\}. For historical input lengths, we follow \citep{FITS} by treating $T$ as a hyperparameter, systematically conducting grid search within \{96, 180, 360, 720\} to identify the optimal length for each model. This approach accounts for observations that some models degrade with longer histories (e.g., iTransformer \citep{itransformer} on ETT datasets \citep{FITS}). Our model (SWIFT) and CycleNet \citep{cyclenet} use a lookback window length of 720 because both of them achieve best performance at $T$=720, as their accuracy improves with extended input lengths. For all other models, we fix other hyperparameters to their original settings from official implementations, except for historical input lengths. We rerun all the experiment with code and script provided by their official implementation.

\paragraph{Evaluation} To avoid information leakage, We choose the hyper-parameter based on the performance of the validation set. We follow the previous works \citep{informer, DLinear, FITS} to compare forecasting performance using Mean Squared Error (MSE) as the core metrics.
\begin{figure}[ht]
  \centering
  \includegraphics[width=0.48\textwidth]{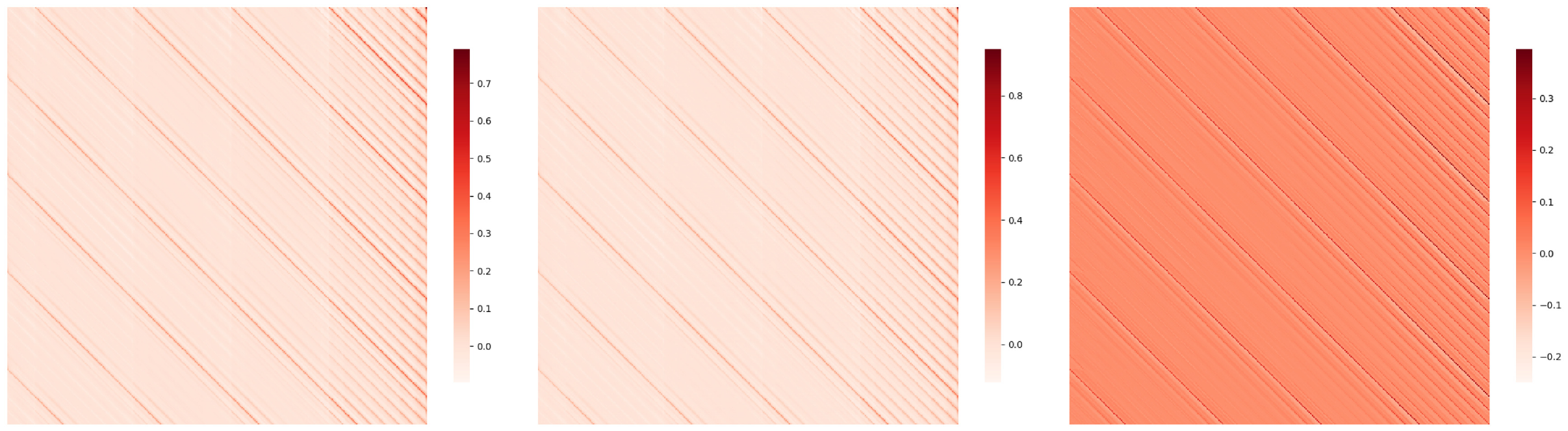}
  \caption{Visualization results of weight maps trained on the ECL dataset. From left to right are \( W_s \), \( W_l \) and \( W_h \).}
  \label{fig:weight}
\end{figure}
\paragraph{Main results} Comprehensive forecasting results are listed in Table \ref{tab:SWIFTett} and Table \ref{tab:SWIFTother} with the best in \textbf{bold} and the second \underline{underlined}. The lower MSE indicates the more accurate prediction result. As shown in table \ref{tab:SWIFTett} and table \ref{tab:SWIFTother}, SWIFT performs well in the forecasting task. Overall, SWIFT achieves state of the art performance. Due to the nonlinear mapping capability of MLP compared to Linear, the MLP version of SWIFT performs better on high-dimensional
datasets such as Electricity and Solar-Energy (i.e., datasets with more than 100 channels). In summary, SWIFT achieves comparable or even superior performance in all 7 datasets with even a very simple Linear or MLP, while requiring over $100\times$ fewer parameters than many existing methods (i.e., Transformer-based methods).

\begin{table}[ht]
\centering
\resizebox{0.48\textwidth}{!}{
\begin{tabular}{c|cccc}
\toprule
Model & Parameters & MACs & Train./epoch (GPU) \\
\midrule
DLinear & 138.4k & 44.61 M & \emph{19.062s} \\
FITS & 116.2k & 1189.91 M & \emph{25.070s} \\
CycleNet / Linear & 123.7k & 22.42M & \emph{28.268s} \\
CycleNet / MLP & 472.9k & 134.84M & \emph{30.200s} \\
\midrule
SWIFT / MLP & 53.1k & 33.53 M & \emph{19.717s} \\ 
SWIFT / Linear & 18.1k & 11.09 M & \emph{18.571s} \\
\bottomrule
\end{tabular}
}
\caption{Number of trainable parameters, MACs, and training time of models with less than 1M parameters, under look-back window=720 and forecasting horizon=96 on Electricity.}
\label{tab:para}
\end{table}

\begin{table*}[htbp]
\centering
\begin{tabular}{@{}c|cccc|cccc|cccc@{}}
\toprule
Dataset & \multicolumn{4}{c|}{ETTh1} & \multicolumn{4}{c|}{ETTm1} & \multicolumn{4}{c}{traffic} \\ \midrule
Horizon & 96 & 192 & 336 & 720 & 96 & 192 & 336 & 720 & 96 & 192 & 336 & 720 \\ 
\midrule
\textbf{SWIFT} & 0.367 & \textbf{0.395} & \textbf{0.420} & \textbf{0.430} & \textbf{0.307} & \textbf{0.336} & \textbf{0.364} & \textbf{0.413} & \textbf{0.368} & \textbf{0.382} & \textbf{0.396} & \textbf{0.430}\\ 
$w/o \ Conv$ & 0.376 & 0.413 & 0.463 & 0.443 & 0.310 & 0.337 & 0.377 & 0.439 & 0.371 & 0.384 & 0.427 & 0.470 \\ 
$w/o \ DWT$ & \textbf{0.365} & 0.399 & 0.427 & 0.446 & 0.321 & 0.338 & 0.365 & 0.414 & 0.521 & 0.674 & 0.533 & 0.717  \\ 
\bottomrule
\end{tabular}
\caption{Ablation study 1 of SWIFT. For dataset etth1 and ettm1, we choose Linear version of SWIFT; for dataset traffic, we use MLP version of SWIFT.}
\label{tab:ablations}
\end{table*}

\begin{table*}[htbp]
\centering
\resizebox{1.0\textwidth}{!}{
\begin{tabular}{@{}c|cccc|cccc|cccc|cccc@{}}
\toprule
Dataset & \multicolumn{4}{c|}{ETTh1} & \multicolumn{4}{c|}{ETTh2} & \multicolumn{4}{c|}{ETTm1} & \multicolumn{4}{c}{ETTm2} \\ \midrule
Horizon & 96 & 192 & 336 & 720 & 96 & 192 & 336 & 720 & 96 & 192 & 336 & 720 & 96 & 192 & 336 & 720 \\ 
\midrule
$Share$ & \textbf{0.367} & \textbf{0.395} & \textbf{0.420} & 0.430 & 0.268 & \textbf{0.329} & \textbf{0.351} & 0.383 & 0.307 & 0.336 & \textbf{0.364} & \textbf{0.413} & \textbf{0.161} & 0.214 & 0.267 & \textbf{0.348} \\ 
$Split$ & 0.368 & \textbf{0.393} & 0.421 & 0.433 & \textbf{0.266} & \textbf{0.329} & 0.353 & \textbf{0.380} & \textbf{0.305} & \textbf{0.334} & 0.366 & 0.414 & 0.162 & 0.215 & \textbf{0.266} & \textbf{0.348} \\ 
\midrule
IMP & 0.001 & -0.002 & 0.001 & 0.003 & -0.002 & 0.000 & 0.002 & -0.
003& -0.002 & -0.002 & 0.002 & 0.001 & 0.001 & 0.001 & -0.001 & 0.000 \\ 
\bottomrule
\end{tabular}
}
\caption{Ablation study 2 of SWIFT. IMP is the improvement between with $Share$ and $Split$ result, where a larger value indicates a better improvement. $Share$ corresponds to the use of only one linear layer, while $Split$ corresponds to the use of one linear layer for each frequency band. The model we used is SWIFT / Linear.}
\label{tab:share}
\end{table*}

Table \ref{tab:para} presents the number of trainable parameters and MACs for various linear-based time series forecasting (TSF) models using a look-back window of 720 and a forecasting horizon of 96 on the Electricity dataset. The table clearly demonstrates the exceptional efficiency of SWIFT compared to other models. Among all efficient models, SWIFT stands out with significantly fewer parameters and much faster training times. SWIFT-Linear requires only 15\% of the FITS parameters and 60\% of its MACs, while achieving comparable or even superior performance to these state-of-the-art efficient forecasting models. It should be noted that SWIFT-Linear’s parameter count is also much lower than Dlinear, which has 138.4K parameters. Moreover, while the parameter count of FITS increases rapidly when using longer look-back windows for forecasting, SWIFT does not exhibit this issue.

\begin{table}[htbp]
\centering
\resizebox{0.48\textwidth}{!}{
\begin{tabular}{@{}l|l|c|c|c|c@{}}
\toprule
Dataset & $Sim_{s,l}$ & $Sim_{s,h}$ & $Sim_{l,h}$ & LR equation & MSE \\
\midrule
ETTh1 & 93.5\% & 24.7\% & 23.1\% & $\mathbf{W_s} \approx 0.8825\mathbf{W_l} + 0.0538\mathbf{W_h} + 0.0018$ & 0.000\\
ETTh2 & 95.0\% & 28.6\% & 20.1\% & $\mathbf{W_s} \approx 0.8201\mathbf{W_l} + 0.0568\mathbf{W_h} + 0.0028$ & 0.000\\
ETTm1 & 97.1\% & 60.5\% & 28.6\% & $\mathbf{W_s} \approx 0.9263\mathbf{W_l} + 0.0197\mathbf{W_h} + 0.0089$ & 0.000\\
ETTm2 & 88.2\% & 42.1\% & 40.6\% & $\mathbf{W_s} \approx 0.6887\mathbf{W_l} + 0.0530\mathbf{W_h} - 0.0001$ & 0.000\\
ECL & 97.3\% & 54.1\% & 51.5\% & $\mathbf{W_s} \approx 0.9119\mathbf{W_l} + 0.0482\mathbf{W_h} - 0.0001$ & 0.000\\
Traffic & 97.0\% & 52.4\% & 47.7\% & $\mathbf{W_s} \approx 0.8679\mathbf{W_l} + 0.0700\mathbf{W_h} + 0.0040$ & 0.000\\
Weather & 94.1\% & -1.0\% & -1.0\% & $\mathbf{W_s} \approx 0.7706\mathbf{W_l} + 0.0017\mathbf{W_h} - 0.0002$ & 0.000\\
\midrule
\end{tabular}
}
\caption{Analysis results between weight matrices. We trained both variants on 7 datasets and analyzed their linear layer weights. We used \textit{cosine similarity} and \textit{linear regression analyses} to explore the relationships that exist between the three weight matrices. $Sim$ denotes the absolute value of \textit{cosine similarity}, LR equation shows a concrete representation between matrices. MSE stands for loss of fit in \textit{linear regression}.}
\label{tab:analysis}
\end{table}
\paragraph{Ablations} 
We conduct two ablation experiments on proposed SWIFT, which are shown in Table \ref{tab:ablations} and Table \ref{tab:share}. 

For the first ablation study, we choose three datasets for the convolutional layer ablation  experiments and DWT module ablation experiments which is shown in Table \ref{tab:ablations}. Obviously, the role of the convolution layer in our model is crucial, which can be proved by the overall increase in performance. In SWIFT, the convolution layer not only denoises sequences and captures timing dependencies, but also enables cross-band feature integration. Besides, the performance without the DWT module is poor, which demonstrates the critical importance of DWT in our model.

Table \ref{tab:share} shows that the high-frequency and low-frequency components obtained after wavelet decomposition are able to share a linear layer for mapping, which does not result in performance loss. From the perspective of Structural Risk Minimization (SRM), parameter sharing constrains the hypothesis space, acting as an implicit regularizer that improves generalization and mitigates overfitting to band-specific fluctuations.
 After conducting an in-depth study, we came to the following two conclusions: (i) The components of different frequency bands obtained after DWT may have some underlying feature correlation, so they can be represented and mapped in the same feature space. (ii) Convolution layer enables cross-band feature fusion. 

To investigate the intrinsic connection between the $Share$ and $Split$ strategies and enhance the interpretability of our conclusions, we conducted an analysis of the linear weight matrices from the two model variants. Specifically, we denote the linear weight matrix of the $Share$ strategy as \( W_s \), and the linear weight matrices of the $Split$ strategy, used for forecasting the low-frequency and high-frequency components, as \( W_l \) and \( W_h \), respectively. We utilized two metrics to represent the relationships among the three matrices: cosine similarity and linear regression (LR). We use the following formula to get the similarity representation:
\[
Sim_{a, b}=\frac{\sum_{i = 1}^{n}a_ib_i}{\sqrt{\sum_{i = 1}^{n}a_i^2}\sqrt{\sum_{i = 1}^{n}b_i^2}}
\]

Since we found that the \( W_s \), \( W_l \) and \( W_h \) have a very high similarity in their pattern, we hypothesized that there is a presentness relationship between the \( W_s \) and \( W_l \) and \( W_h \):
\[
W_s = \beta_{l} W_l + \beta_{h} W_h + \epsilon
\]
where \( W_s \), \( W_l \) and \( W_h \) $\in \mathbb{R}^{T/2 \times T/2}$. Then we fit $\beta_{l}, \beta_{h}$ and $\epsilon$ using LR. Both metrics were implemented using the machine learning library scikit-learn \citep{scikit-learn}.

To obtain trained model weights, both variants were trained on seven datasets using identical hyperparameter settings and a sequence length of 720-720 for 10 epochs. The weights corresponding to the best performance on the validation set were saved for further analysis. Experiment results and visualization results are listed in Table \ref{tab:analysis} and Figure \ref{fig:weight}.

The experimental results demonstrate that the three matrices can be accurately regressed using a simple LR model, with reconstruction MSE nearly zero (up to three decimal places). This indicates that the shared strategy inherently learns a linear combination pattern of high-frequency and low-frequency components. Despite their seemingly distinct patterns, these components can be effectively represented using shared parameters. This strategy not only reduces the number of parameters but also preserves performance, enabling information exchange between the components in a combinatorial manner. Additionally, the \( W_s \) exhibits a very high similarity to the \( W_l \), suggesting that in the wavelet domain, the low-frequency components play a dominant role in prediction. Meanwhile, the high-frequency components, which contain significant levels of noise and various non-stationary elements, contribute only marginally to the final prediction outcomes, yet still play a small but notable role in the overall forecasting process.

\section{Conclusion and Future work}

In this paper, we propose SWIFT for time series analysis, an efficient model with performance comparable to state-of-the-art models that are typically several orders of magnitude larger. In future work, we plan to evaluate SWIFT in a broader range of real-world scenarios, including but not limited to anomaly detection and classification tasks, to validate its robustness and generalizability across diverse applications. In addition, we plan to explore large neural networks in the time-frequency domain to perform scaling up operations on SWIFT and improve its prediction performance. We also intend to further investigate multi-resolution wavelet transforms to better leverage multi-scale information for time series forecasting. This exploration could lead to more robust and adaptive representations in complex and non-stationary environments.

\section*{Impact Statement}

This paper presents SWIFT, a lightweight and efficient model designed to advance the field of Long-term Time Series Forecasting (LTSF). By significantly reducing the number of parameters and computational overhead, our work enables high-performance forecasting to be deployed on resource-constrained edge devices. This contributes to the democratization of AI by making advanced predictive capabilities accessible for latency-sensitive and low-power applications, such as energy scheduling and intelligent transportation systems. While there are many potential societal consequences of improving forecasting efficiency, including reduced energy consumption for AI inference, there are no specific ethical concerns or negative consequences that we feel must be specifically highlighted here.

\nocite{langley00}

\bibliography{example_paper}
\bibliographystyle{icml2026}
\newpage
\appendix
\onecolumn
\section{Theoretical Justification}

In this section, we provide a theoretical foundation demonstrating that a Multilayer Perceptron (MLP) trained on a finite window of univariate history can effectively model the dynamics of a coupled system.

\begin{proposition}
Let $\mathcal{S}$ be a dynamical system governing an observable variable $y_t$ and a latent state $x_t \in \mathbb{R}^d$.
Under the assumption of fading memory, the future observation $y_{t+1}$ can be approximated by a function of a finite history window $y_{t:t-k}$.
Specifically, the approximation error is bounded by a term that decays exponentially with the window size $k$, regardless of whether the strict topological reconstruction condition (Takens' Embedding Theorem) is fully met.
\end{proposition}

\begin{proof}
Consider the coupled discrete-time system:
\begin{align}
    y_{t+1} &= f(y_t, x_t) \\
    x_{t+1} &= g(x_t, y_t)
\end{align}
where $f$ and $g$ are smooth functions, and the system evolves on a compact manifold $\mathcal{M}$ of dimension $D$.

\paragraph{Step 1: Existence of the Mapping (Takens' Theorem).}
According to Takens' Embedding Theorem (Takens, 1981), for a generic smooth observable $y$, the delay coordinate map $\Phi: \mathcal{M} \to \mathbb{R}^{m}$ defined by $\Phi(s_t) = [y_t, y_{t-1}, \dots, y_{t-(m-1)}]$ is a diffeomorphism, provided the embedding dimension (history window) satisfies $m > 2D$.
This implies that the delay vector is topologically equivalent to the full state $s_t = (y_t, x_t)$. Consequently, there exists a deterministic function $\mathcal{F}$ such that:
\begin{equation}
    y_{t+1} = \mathcal{F}(y_t, y_{t-1}, \dots, y_{t-m+1})
\end{equation}

\noindent\textbf{Remark on Dimensionality.}
For most datasets considered (e.g., Weather, Electricity, Exchange), the effective dimension $D$ of the underlying system is relatively low. Thus, a standard look-back window $k$ (e.g., $k=96$ or $k=336$) typically satisfies the condition $k > 2D$, ensuring the theoretical reconstruction of the phase space.
However, a notable exception is the \textit{Traffic} dataset. The traffic system involves complex spatiotemporal interactions with extremely high degrees of freedom ($D_{\text{traffic}} > k$), meaning the condition $k > 2D$ may not be strictly met. In such high-dimensional cases, we rely on the \textit{Fading Memory} property (Step 2) and the \textit{Universal Approximation} capability (Step 3) to approximate the dynamics locally, rather than guaranteeing a global topological reconstruction.

\paragraph{Step 2: Fading Memory and Remainder Analysis.}
We relax the strict reconstruction requirement by utilizing the \textit{Fading Memory} assumption. By recursively expanding the state update equation $x_t = g(x_{t-1}, y_{t-1})$ for $k$ steps, we express $y_{t+1}$ as:
\begin{equation}
    y_{t+1} = G(\mathbf{y}_{t:t-k}) + R_k(x_{t-k})
\end{equation}
Assuming the system is Lipschitz continuous with a contraction factor $\lambda < 1$, the remainder term is bounded by:
\begin{equation}
    |R_k| \le C \cdot \lambda^k \cdot \|x_{t-k}\|
\end{equation}
As $k$ increases, the influence of the unobserved initial state $x_{t-k}$ decays exponentially ($|R_k| \to 0$). This ensures that even if the embedding dimension is insufficient to fully unfold the attractor (as in Traffic data), the recent history $\mathbf{y}_{t:t-k}$ still dominates the prediction of $y_{t+1}$.

\paragraph{Step 3: Universal Approximation.}
Since the derived mapping $G$ is continuous, by the Universal Approximation Theorem (Hornik et al., 1989), there exists an MLP parameterized by $\theta$ such that for any $\epsilon > 0$:
\begin{equation}
    \sup | G(\mathbf{y}_{t:t-k}) - \text{MLP}_\theta(\mathbf{y}_{t:t-k}) | < \epsilon
\end{equation}
Combining these steps, the total approximation error is bounded by:
\begin{equation}
    | y_{t+1} - \text{MLP}_\theta(\mathbf{y}_{t:t-k}) | \le \epsilon + \mathcal{O}(\lambda^k)
\end{equation}
This completes the justification.
\end{proof}

\section{Experimental Details} \label{experimental_details}
\subsection{Datasets} \label{appendix_datasets}
We evaluate the performance of our proposed SWIFT on 7 popular datasets, including Weather, Traffic, Electricity, and ETT datasets. The Weather dataset, including 21 meteorological indicators such as air temperature and humidity, is collected every 10 minutes from the Weather Station of the Max Planck Institute for Biogeochemistry in 2020. The Traffic dataset contains hourly traffic data measured by 862 sensors on San Francisco Bay area freeways, which has been collected since January 1, 2015. The Electricity dataset collects the hourly electricity consumption of 321 clients from 2012 to 2014. The ETT (Electricity Transformer Temperature) datasets contain two visions of the sub-dataset: ETTh and ETTm, collected from electricity transformers every 15 minutes and 1 hour between July 2016 and July 2018. Thus, in total we have 4 ETT datasets (ETTm1, ETTm2, ETTh1, and ETTh2) recording 7 features such as load and oil temperature. The details about these datasets are summarized in Table \ref{tab:dataset}.
\begin{table*}[ht]
    \centering
    \scalebox{0.9}{
    \begin{tabular}{c c|c c c c c c c c}
    \toprule[1pt]
    \toprule
       \multicolumn{2}{c|}{Datasets}  &ETTh1  &ETTh2  &ETTm1  &ETTm2   &Electricity  &Weather &Traffic &Exchange \\
       \midrule
       \midrule
       \multicolumn{2}{c|}{Variables} &7      &7      &7       &7       &321          &21        &862      &8 \\
       \multicolumn{2}{c|}{Timesteps} &17420  &17420  &69680   &69680   &26304        &52696     &17544    &7588 \\
       \multicolumn{2}{c|}{Frequency} &Hourly &Hourly &15min   &15min   &Hourly       &10min     &Hourly   &Daily \\
       \multicolumn{2}{c|}{Information} &Electricity &Electricity &Electricity &Electricity &Electricity &Weather &Traffic &Economy \\
    \bottomrule[1pt]
    \bottomrule
    \end{tabular}
    }
    \caption{The details of datasets.}
    \label{tab:dataset}
\end{table*}

\subsection{Baselines} \label{appendix_baselines}

We choose several well-acknowledged and state-of-the-art models for comparison to evaluate the effectiveness of our proposed SWIFT for time series forecasting, including MLP-based models and Transformer-based models. We introduce these models as follows:

\textbf{FITS}~\cite{FITS} performs time series analysis through interpolation in the complex frequency domain, enjoying low cost with 10k parameters.

\textbf{TimesNet}~\cite{timesnet} transforms 1D time series into a set of 2D tensors based on multiple periods to analyse temporal variations. The above transformation allows the 2D-variations to be easily captured by 2D kernels with encoding the intraperiod- and interperiod-variations into the columns and rows of the 2D tensors respectively.

\textbf{DLinear}~\cite{DLinear} utilizes a simple yet effective one-layer linear model to capture temporal relationships between input and output sequences.

\textbf{Informer}~\cite{informer} enhances Transformer with KL-divergence based ProbSparse attention for $O(L\operatorname{log}L)$ complexity, efficiently encoding dependencies among variables and introducing a novel architecture with a DMS forecasting strategy.

\textbf{FEDformer}~\cite{fedformer} implements sparse attention with low-rank approximation in frequency domain, enjoying linear computational complexity and memory cost. And it proposes mixture of experts decomposition to control the distribution shifting.

\textbf{PatchTST}~\cite{patch} divides time series data into subseries-level patches to extract local semantic information and adopts channel-independence strategy where each channel shares the same embedding and Transformer weights across all the series.

\textbf{iTransformer}~\cite{itransformer} inverts the structure of Transformer without modifying any existing modules by encoding
individual series into variate tokens. These tokens are utilized by the attention mechanism to capture multivariate correlations and FFNs are adopted for each variate token to learn nonlinear representations.

\section{Visualizations}

\begin{figure}[htbp]
  \centering  
  \includegraphics[width=0.75\textwidth]{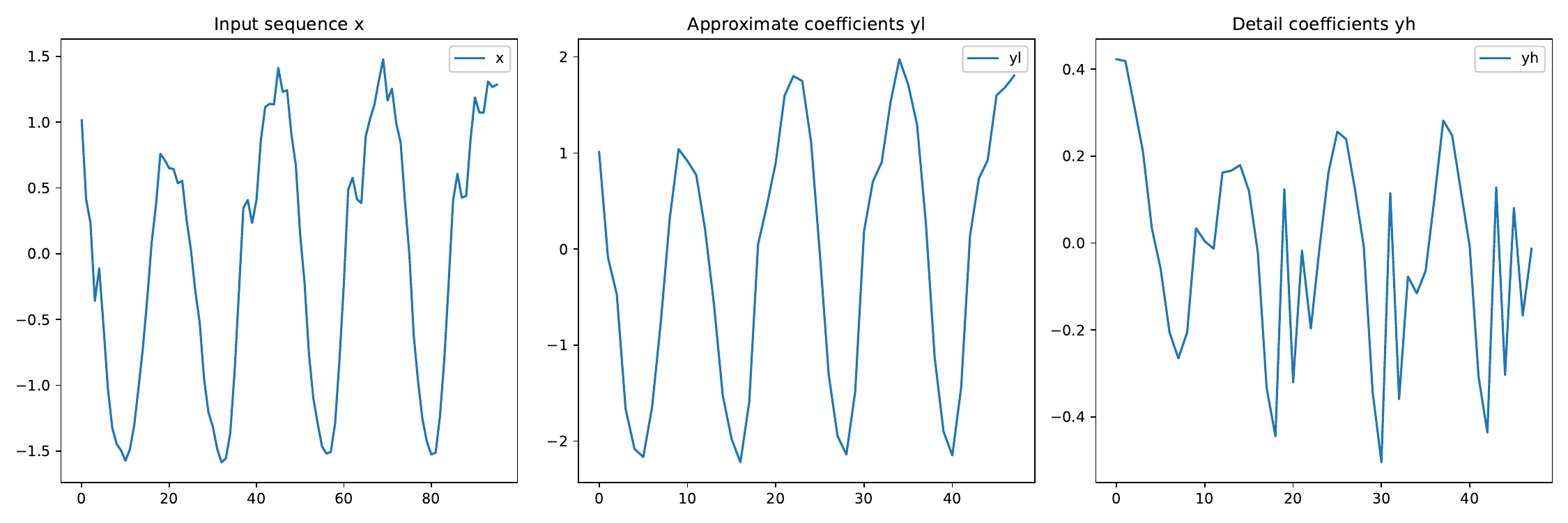}
  \caption{Visualization of DWT decomposition on the Traffic.}
  \label{fig:DWT visualize_1}
  \includegraphics[width=0.75\textwidth]{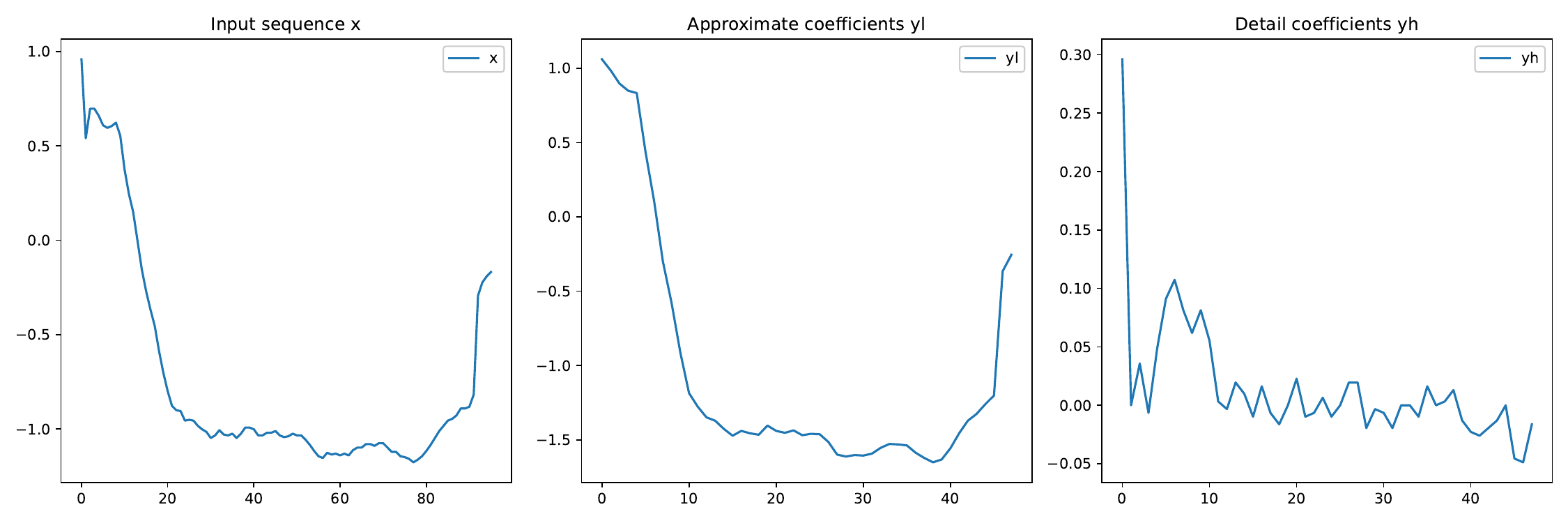}
  \caption{Visualization of DWT decomposition on the Weather.}
  \label{fig:DWT visualize_2}
  \includegraphics[width=0.75\textwidth]{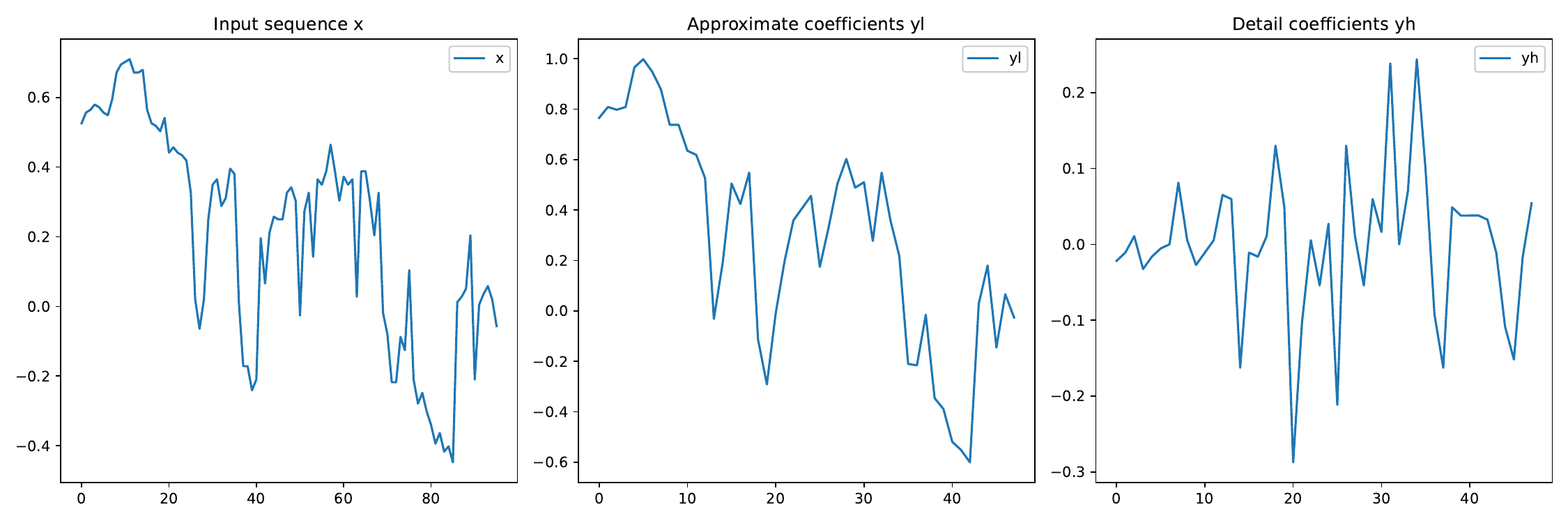}
  \caption{Visualization of DWT decomposition on the ETT.}
  \label{fig:DWT visualize_3}
\end{figure}

We applied Discrete Wavelet Transform (DWT) using the Haar wavelet to three datasets, decomposing each into low-frequency (approximation) and high-frequency (detail) components, as shown in Figure \ref{fig:DWT visualize_1}, Figure \ref{fig:DWT visualize_2} and Figure \ref{fig:DWT visualize_3}. The low-frequency component reduces data length by half, enhancing processing efficiency while preserving key patterns like trends and structural features. The high-frequency component captures local fluctuations and fine-grained variations, crucial for detailed data analysis.

DWT enables efficient data compression and simplification while retaining core information. The low-frequency component optimizes the SWIFT model's inference efficiency, reducing parameters and computational complexity, while the high-frequency component ensures detailed information integrity, supporting predictive performance. This approach provides a robust foundation for time series analysis and prediction, balancing computational efficiency and information preservation for real-time applications.

\section{Anomaly detection results}

\paragraph{Datasets} We conducted experiments on 4 benchmark datasets used by \citep{anomaly}: SMD (Server Machine Dataset), PSM (Polled Server Metrics), MSL (Mars Science Laboratory rover) and SMAP (Soil Moisture Active Passive satellite).
\paragraph{Baseline} We compare SWIFT with models such as TimesNet \citep{timesnet}, Anomaly Transformer \citep{anomaly}, THOC \citep{THOC}, Omnianomaly \citep{Omnianomaly}, DGHL \citep{DGHL}. Following TimesNet, we also compare the anomaly detection performance with other models \citep{DLinear}.
\paragraph{Main results} Table \ref{tab:AD} demonstrates the performance of SWIFT on various datasets. SWIFT achieves near-perfect F1 scores of 99.92\% and 96.487\% on the SMD and PSM datasets, respectively, and achieves about 2.5\% outperformance on PSM compared to FITS, demonstrating its high-precision performance in anomaly detection. By introducing the time-frequency domain, SWIFT can easily identify the anomalies that are difficult to recognize in the time domain, including the anomalies that introduce unexpected time-frequency components. In contrast, models such as TimesNet, Anomaly Transformer and Stationary Transformer do not perform as well as SWIFT on these datasets.

However, SWIFT performs relatively low on the SMAP and MSL datasets. Due to the binary event data nature of these datasets, time-frequency feature representations may have difficulty capturing these features effectively. In this case, time-domain modeling is more desirable because the raw data format is already compact enough. As a result, models designed for anomaly detection, such as THOC and Omni Anomaly, achieve higher F1 scores on these datasets.

It is worth noting that SWIFT has an extremely low number of parameters ranging from 0.625k-2.5k for the anomaly detection task. This feature allows SWIFT to be deployed on almost any edge device. In addition, SWIFT's inference speed is impressive, reaching sub-millisecond levels, much faster than the latency associated with larger models or communication overheads. This speed underscores SWIFT's suitability as a first-response tool for rapid detection of critical errors.
\begin{table*}[t]
\centering
\resizebox{\textwidth}{!}{
        \begin{tabular}{@{}cccccccccccccc@{}}
        \toprule
        Models & SWIFT  & FITS  & TimesNet  & Anomaly & THOC & Omni & Stationary & DGHL & OCSVM & IForest & LightTS & Dlinear\\ 
        \midrule
        \multicolumn{1}{c|}{SMD}  & \underline{99.92} & \textbf{99.95} & 85.81 & 92.33 & 84.99 & 85.22 & 84.72 & N/A   & 56.19 & 53.64 & 82.53  & 77.1\\
        \multicolumn{1}{c|}{PSM}  & 96.47  & 93.96 & 97.47 & \underline{97.89} & \textbf{98.54} & 80.83 & 97.29 & N/A   & 70.67 & 83.48 & 97.15  & 93.55 \\
        \multicolumn{1}{c|}{SMAP} & 68.16 & 70.74 & 71.52 & \textbf{96.69} & 90.68 & 86.92 & 71.09 & \underline{96.38} & 56.34 & 55.53 & 69.21  & 69.26\\
        \multicolumn{1}{c|}{MSL}  & 58.51 & 78.12 & 85.15 & \underline{93.59}  & 89.69 & 87.67 & 77.5  & \textbf{94.08} & 70.82 & 66.45 & 78.95  & 84.88 \\ 
        \bottomrule
        \end{tabular}
    }
    \caption{Anomaly detection result of F1-scores on 4 datasets. The best result is highlighted in \textbf{bold}, and the second best is highlighted with \underline{underline}.}
\label{tab:AD}
\end{table*}
\section{Additional experiments}

\subsection{Ablations on Channel Independence}

To investigate the impact of Channel Independence (CI) on model performance, we conduct an ablation study comparing SWIFT models with and without the CI mechanism across four ETT benchmark datasets and Traffic dataset. Table \ref{tab:CI_comparision} presents the Mean Squared Error (MSE) results for different prediction horizons.

\begin{table}[htbp]
\centering
\begin{threeparttable}
\setlength{\tabcolsep}{10pt} 
\renewcommand{\arraystretch}{1.2} 
\begin{tabular}{l|c|c|c|c|c}
\toprule
\textbf{Dataset} & \textbf{Model} & \textbf{L=96} & \textbf{L=192} & \textbf{L=336} & \textbf{L=720} \\
\hline
\multirow{2}{*}{ETTh1} 
& Not $CI$ & \textbf{0.367} & \textbf{0.395} & \textbf{0.420} & \textbf{0.430} \\
& $CI$ & 0.375 & 0.413 & 0.430 & 0.450 \\
\hline
\multirow{2}{*}{ETTh2} 
& Not $CI$ & \textbf{0.268} & \textbf{0.329} & \textbf{0.351} & \textbf{0.383} \\
& $CI$ & 0.293 & 0.337 & 0.365 & 0.398 \\
\hline
\multirow{2}{*}{ETTm1} 
& Not $CI$ & 0.307 & 0.336 & \textbf{0.364} & \textbf{0.413} \\
& $CI$ & \textbf{0.300} & \textbf{0.335} & 0.368 & 0.420 \\
\hline
\multirow{2}{*}{ETTm2} 
& Not $CI$ & \underline{0.161} & \textbf{0.214} & \textbf{0.267} & \underline{0.348} \\
& $CI$ & 0.161 & 0.215 & 0.268 & 0.348 \\
\hline
\multirow{2}{*}{Traffic} 
& Not $CI$ & \textbf{0.368} & \textbf{0.382} & \textbf{0.396} & \textbf{0.430} \\
& $CI$ & 0.425 & 0.444 & 0.462 & 0.487 \\
\bottomrule
\end{tabular}
\begin{center}
    \footnotesize
    Note: All experiments use sequence length = 720 and seed = 2023.\\
    \textbf{Bold}: Best performance. \underline{underline}: Performance of the tie.
\end{center}

\end{threeparttable}
\caption{Ablation results on Channel Independence in SWIFT Model Performance (MSE) across ETT and Traffic.}
\label{tab:CI_comparision}
\end{table}

\subsection{Ablations on Wavelets}

To investigate the impact of different wavelet filters on SWIFT model performance, we conduct a comprehensive ablation study comparing three representative wavelets: Haar, Daubechies-2 (DB2), and Symlet-4 (Sym4) across multiple benchmark datasets. The choice of wavelet filter is crucial for the SWIFT architecture, as it directly affects both computational efficiency and forecasting accuracy.

\begin{table}[htbp]
\centering
\begin{threeparttable}
\setlength{\tabcolsep}{10pt} 
\renewcommand{\arraystretch}{1.2} 
\begin{tabular}{l|c|c|c|c|c}
\hline
\textbf{Dataset} & \textbf{Wavelet} & \textbf{L=96} & \textbf{L=192} & \textbf{L=336} & \textbf{L=720} \\
\hline
\multirow{4}{*}{ETTh1} 
& Haar & \textbf{0.367} & \textbf{0.395} & \textbf{0.420} & \textbf{0.430} \\
& DB2 & 0.377 & 0.413 & 0.444 & 0.456 \\
& Sym4 & 0.374 & 0.412 & 0.447 & 0.462 \\
& Coif & 0.375 & 0.430 & 0.450 & 0.459 \\
\hline
\multirow{4}{*}{ETTm1} 
& Haar & \textbf{0.307} & \textbf{0.336} & \textbf{0.364} & \textbf{0.413} \\
& DB2 & 0.309 & 0.343 & 0.373 & 0.418 \\
& Sym4 & 0.310 & 0.339 & 0.365 & 0.415 \\
& Coif & 0.320 & 0.341 & 0.368 & 0.415 \\
\hline
\multirow{4}{*}{Weather} 
& Haar & \textbf{0.140} & \textbf{0.183} & \textbf{0.235} & \textbf{0.307} \\
& DB2 & 0.153 & 0.194 & 0.245 & 0.315 \\
& Sym4 & 0.151 & 0.194 & 0.244 & 0.313 \\
& Coif & 0.152 & 0.195 & 0.243 & 0.315 \\
\hline
\end{tabular}
\begin{center}
    \footnotesize
    Note: All experiments use sequence length = 720 and seed = 2023.\\
    \textbf{Bold}: Best performance. \underline{underline}: Performance of the tie.
\end{center}

\end{threeparttable}
\caption{Ablation Study on Different Wavelets in SWIFT Model Performance (MSE).}
\label{tab:ab_result_wavelets}
\end{table}

Table \ref{tab:ab_result_wavelets} presents the comparative results of MSE performance across different prediction horizons. The experimental findings reveal that Haar wavelets consistently achieve the best performance across all datasets and prediction lengths, demonstrating universally superior forecasting accuracy. Specifically, Haar wavelets outperform both DB2 and Sym4 wavelets with MSE improvements ranging from 2.7\% to 5.7\% on ETTh1 compared to the second-best performer. On ETTm1, Haar maintains consistent advantages with improvements of 0.6\% to 2.8\% over alternative wavelets.

Most notably, the Weather dataset shows the most significant performance gains with Haar wavelets, achieving improvements of 7.3-8.5\% (L=96), 5.7\% (L=192), 3.7-4.1\% (L=336), and 1.9-2.5\% (L=720) compared to DB2 and Sym4. Interestingly, DB2 and Sym4 wavelets exhibit similar performance patterns, with Sym4 showing marginal improvements over DB2 in some configurations, but neither approaching the consistent superiority of Haar wavelets across all experimental scenarios.

Beyond forecasting accuracy, the computational efficiency considerations strongly favor Haar wavelets. The Haar wavelet filter possesses the shortest possible support length among orthogonal wavelets, consisting of only two coefficients [1, -1]. This minimal filter length is particularly crucial for our proposed sharing strategy, which constitutes the core mechanism enabling SWIFT's computational efficiency. Longer wavelet filters, such as DB2 with its four-coefficient structure and Sym4 with its eight-coefficient structure, introduce additional computational overhead and memory requirements that are incompatible with our efficient sharing framework.
The combination of superior forecasting performance and optimal computational characteristics makes Haar wavelets the ideal choice for the SWIFT architecture. The results demonstrate that Haar filters not only maintain competitive accuracy but also provide the necessary efficiency gains that enable real-time processing capabilities essential for practical time series forecasting applications.
\newpage
\section{Implementation details}
We built a simulated non-stationary dataset and found that existing models cannot perform well in predicting non-stationary series. We give the code to build the Dataset here.
\begin{algorithm*}
\caption{Non-Stationary Time Series Generation}
\label{alg:dataset}
\begin{algorithmic}[1]
\REQUIRE $L_{in}, L_{out}, F, A, T_{len}, is\_test$
\ENSURE $Input\_Sequences, Target\_Sequences$

\STATE \textbf{Function} GenerateSequence:
\STATE \quad $t \gets \text{linspace}(0, 2, T_{len})$
\STATE \quad $amp \gets \text{Uniform}(A)$, $\phi \gets \text{Uniform}(0, 2\pi, |F|)$
\STATE \quad $S_{base} \gets \sum_{f, \phi} amp \cdot \sin(2\pi f t + \phi)$
    
\IF{$is\_test$}
    \STATE \quad $S_{non\_stat} \gets \sum_{f, \phi} amp \cdot \sin(2\pi f t^2 + \phi)$
\ELSE
    \STATE \quad $S_{non\_stat} \gets \sum_{f, \phi} amp \cdot \sin(2\pi f (t + 0.2\sin(0.5t)) + \phi)$
\ENDIF
    
\STATE \quad $S_{trend} \gets \sin(0.5t) + 0.2t$
\STATE \quad $\epsilon \gets 0.2 \cdot \mathcal{N}(0, 1)$
\STATE \quad $S_{total} \gets S_{base} + S_{non\_stat} + S_{trend} + \epsilon$
\STATE \quad \textbf{return} Normalize($S_{total}$)

\STATE
\STATE \textbf{Function} GetItem($idx$):
\STATE \quad $X \gets S_{total}[idx : idx + L_{in}]$
\STATE \quad $Y \gets S_{total}[idx + L_{in} : idx + L_{in} + L_{out}]$
\STATE \quad \textbf{return} $(X, Y)$
\end{algorithmic}
\end{algorithm*}

\end{document}